\documentclass[10pt,twocolumn,letterpaper]{article}

\usepackage{cvpr}
\usepackage{times}
\usepackage{epsfig}
\usepackage{graphicx}
\usepackage{amsmath}
\usepackage{amssymb}

\usepackage[pagebackref=true,breaklinks=true,colorlinks,bookmarks=false]{hyperref}

\cvprfinalcopy 

\def\cvprPaperID{1412} 

\ifcvprfinal\fi
\begin{document}

\setcounter{footnote}{-1}
\title{Comparative Deep Learning of Hybrid Representations\\
for Image Recommendations
\thanks{This work was supported
by the National Program on Key Basic Research Projects (973 Program)
under Grant 2015CB351800, by the Natural Science Foundation of China
(NSFC) under Grants 61303149, 61331017, 61390512, and 61472392, and
by the Fundamental Research Funds for the Central Universities under Grants
WK2100060011 and WK3490000001.}
}

\author{Chenyi Lei, Dong Liu, Weiping Li, Zheng-Jun Zha, Houqiang Li\\
CAS Key Laboratory of Technology in Geo-Spatial Information Processing and Application System,\\
University of Science and Technology of China, Hefei 230027, China\\
{\tt\small leichy@mail.ustc.edu.cn, \{dongeliu,wpli,zhazj,lihq\}@ustc.edu.cn}
}

\maketitle
\thispagestyle{empty}

\begin{abstract}
In many image-related tasks, learning expressive and discriminative representations of images is essential, and deep learning has been studied for automating the learning of such representations. Some user-centric tasks, such as image recommendations, call for effective representations of not only images but also preferences and intents of users over images. Such representations are termed \emph{hybrid} and addressed via a deep learning approach in this paper. We design a dual-net deep network, in which the two sub-networks map input images and preferences of users into a same latent semantic space, and then the distances between images and users in the latent space are calculated to make decisions. We further propose a comparative deep learning (CDL) method to train the deep network, using a pair of images compared against one user to learn the pattern of their relative distances. The CDL embraces much more training data than naive deep learning, and thus achieves superior performance than the latter, with no cost of increasing network complexity. Experimental results with real-world data sets for image recommendations have shown the proposed dual-net network and CDL greatly outperform other state-of-the-art image recommendation solutions.
\end{abstract}

\section{Introduction}
With the increasing abundance of images, finding out images that satisfy user needs from a huge collection is more and more required, which emphasizes the importance of image search and image recommendations working as filters for users. Such tasks are not trivial, however, due to the gap in understanding the semantics of images as well as the gap in understanding the intents or preferences of users over images. Compared to their counterparts for structured data, such as search of text and recommendations of book or movie, image search and recommendations raise more challenges since images lack an immediately effective representation.

How to represent images both expressively and discriminatively is of essential importance in many image-related tasks including detection, registration, recognition, classification, and retrieval. This problem had been extensively studied, and many kinds of hand-crafted features had been designed and adopted in different tasks~\cite{HOG,SIFT,LLC}. Most of previous work focuses on low-level visual features of images, but for image search and recommendations, it is often not clear how to represent the intents or preferences of users within the framework of low-level features.

One feasible solution that has been studied is to utilize the users' information as constraints to refine the image representations, making them consistent with both semantic labels and user provided hints~\cite{SIDL,Pan,LatentFea}. For example, Liu \etal~\cite{SIDL} proposes to learn an image distance metric by combining the images' visual similarity and their ``social similarity,'' defined from users' interests in images that are mined from user data in online social networks. Nonetheless, visual content of images and users' intents/preferences on images are of two different modalities, simply combining them may not turn out efficient enough.

Recently, deep network models have attracted much attention of researchers in the image processing field. One significant advantage of deep networks is the \emph{automated} learning of image representations, which are demonstrated to be more effective than hand-crafted features, especially in semantic level image understanding~\cite{AlexNet}. Moreover, deep networks have achieved great success in processing other forms of data such as speech and text~\cite{w2v}. Promisingly, multimodal data, such as images and users' intents/preferences, may be efficiently handled by a single integrated deep network.

In this paper, we study a dual-net deep network model for the purpose of making recommendations of images to users. The network consists of two sub-networks, which map an image and the preferences of a user into a same latent semantic space, respectively. Therefore, the network achieves representations of both images and users, termed \emph{hybrid} representations hereafter, and these hybrid representations are directly comparable to make decisions of recommendations.

\begin{figure}[t]
\centerline{\includegraphics[width=\columnwidth]{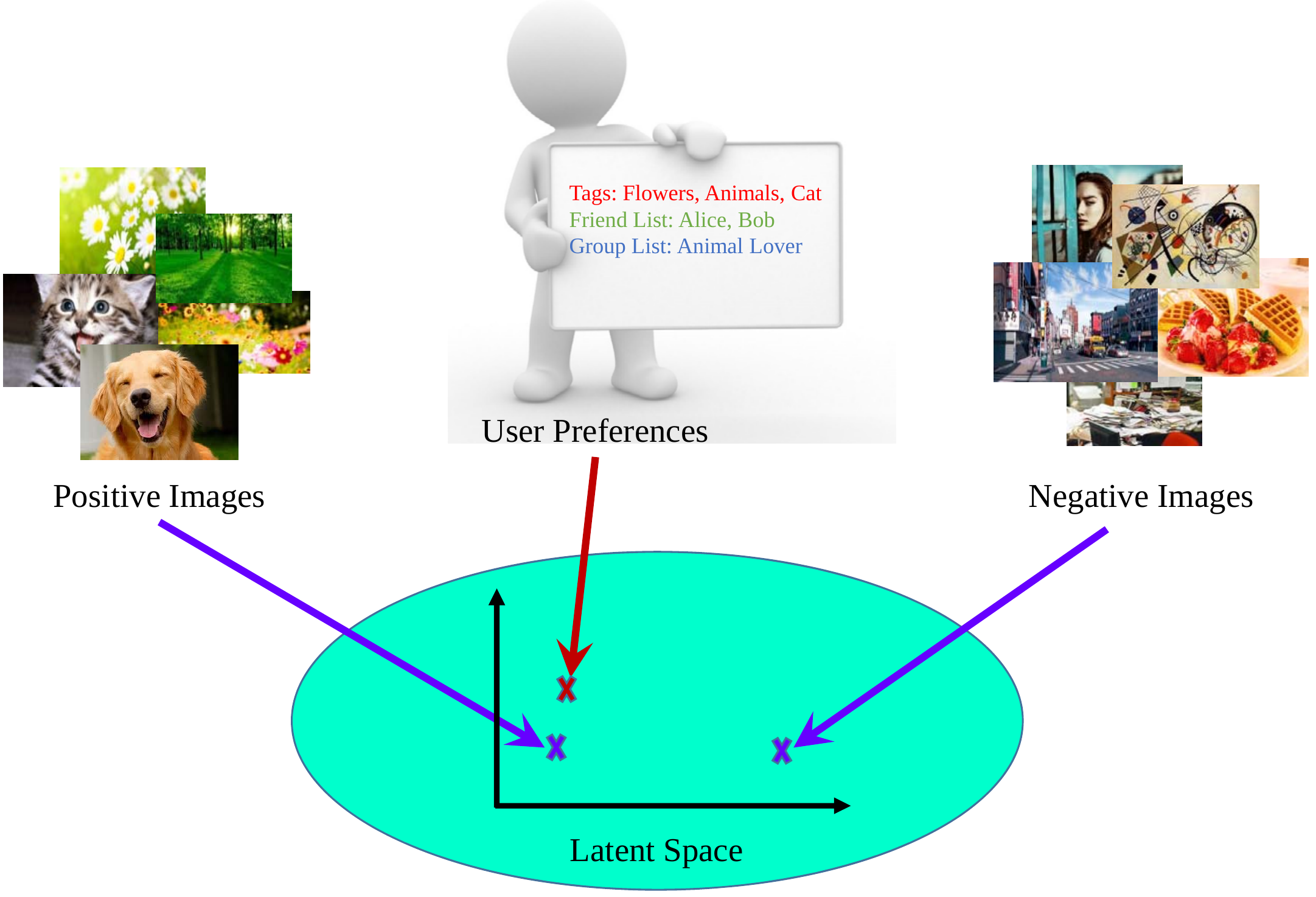}}
\caption{This figure depicts the key idea of our proposed comparative deep learning (CDL). One user's preferences can be described by her frequently used tags as well as her friends' preferences and her joined groups' preferences. These preferences, together with images, are mapped into a same latent semantic space. In that space, the distance between the user and a ``positive'' image (\eg favorite image) shall be less than the distance between the user and a ``negative'' image (\eg disliked image), which is taken as the objective of CDL.}\label{f1}
\end{figure}

Moreover, we propose a comparative deep learning (CDL) method to train the designed deep network. Instead of a naive learning, \eg learning a distance between a user and an image, the CDL uses two images compared against one user, and learns the relative distances among them. Our key idea is depicted in Fig.~\ref{f1}, where for a query user, her historical data used for learning consist of ``positive'' images, \eg her favorites, and ``negative'' images, \eg her dislikes; the objective of CDL is that the distance between the user and a positive image shall be less than the distance between the user and a negative image. Thus, training data for CDL are triplets of $(user, positive\_image, negative\_image)$ and these data are fed into a triple-net deep network consisting of three sub-networks, one of which is for user, and the other two are for positive and negative images and are actually identical, as shown in Fig.~\ref{f2}. Note that after training, we need only two sub-networks for user and image, respectively.

The designed dual-net network and CDL method have been verified on an image recommendation task with real-world data sets. Experimental results display that the proposed CDL achieves superior performance than naive learning, and our proposed solution outperforms other state-of-the-art image recommendation methods significantly.

The remainder of this paper is organized as follows. Related work is discussed in Section 2. Then our proposed CDL-based image recommendation solution is described, the objective of CDL is formulated in Section 3, followed by detailed description of the deep network model in Section 4, and details of making image recommendations in Section 5. Experimental results are reported in Section 6, and concluding remarks in Section 7.

\section{Related Work}
We give brief overview of related work at two aspects: learning of image representations and personalized image recommendations.

\subsection{Learning of Image Representations}

In view of the limitation of hand-crafted image features such as those designed in~\cite{HOG,repreExam1,SIFT,LLC}, more and more research focuses on designing effective deep learning models to extract image representations automatically~\cite{repreExam5,repreExam6,repreExam4}. Karpathy \etal~\cite{repreExam4} proposes a supervised hashing method with deep learning architecture, followed by a stage of simultaneous learning of hash function and image representations. Furthermore, it is noticed that middle-layer outputs in deep learning models can be seamlessly utilized as image representations, though the deep network is not trained for that~\cite{AlexNet,repreExam3,repreExam2}. For example, Krizhevsky \etal~\cite{AlexNet} proposes a deep learning architecture to perform image classification, and the outputs of the 7th full-connection layer are also verified to be kind of robust image representations.

The abovementioned work mainly focuses on low-level visual features of images. But recently, along with the development of user-centric applications such as image recommendations, it is worthwhile to learn not only visual information but also intents or preferences of users for image representations. A paucity of work has made attempts at this aspect~\cite{Hua,SIDL,Pan}. Pan \etal~\cite{Pan} proposes an embedding method to study the cross-view (\ie text to image views) search problem with analyses of user click log. Liu \etal~\cite{SIDL} consider jointly the users' social relationship and images' visual similarity to learn a new image distance metric. But such work relies heavily on carefully designed hand-crafted features. Liu \etal~\cite{NewSIDL} employ deep learning architecture to capture user intent and image visual information, where user intent is described by only similarity between a pair of users. But in practice, there is multi-modal information for drawing upon user intents, such as tags, browsing history and social groups. Moreover, the deep architecture in~\cite{NewSIDL} considers only one image at each training round. To the contrary, recent studies~\cite{Sim,FineGain,Tri} indicate that deep ranking models perform much better by forming training data as triplets. To summary, how to design an effective deep learning architecture to capture both visual information and the intents or preferences of users over images is still a challenging open problem.

\subsection{Personalized Image Recommendations}

Personalized recommendations for structured data such as book, movie, and music have been studied for a long while~\cite{adomavicius2005toward}. Typical technologies include content-based filtering, collaborative filtering, and hybrid of both~\cite{ContentFilter}. However, it is difficult to directly adopt these technologies for image recommendations, possibly due to several difficulties: images are highly unstructured and lack an immediate representations, user-image interaction data are often too sparse, users rarely provide ratings on images but rather give implicit feedback. Nevertheless, mature technologies in recommender systems are still inspiring, for example, matrix factorization~\cite{MF} can be perceived as to learn latent representations of users and items in a same semantic space.

With the development of social networks, recent research starts to leverage social data to improve the performance of recommendations~\cite{SocialRec2,SocialRec1}. Most of existing work on image recommendations also follows this line~\cite{TraRec,TraRec3,TraRec2,RecIcme}. For example, Jing \etal~\cite{TraRec2} propose a novel probabilistic matrix factorization framework that combines the ratings of local community users for recommending Flickr photos. Cui \etal~\cite{TraRec} propose a regularized dual-factor regression method based on matrix factorization to capture the social attributes for recommendations. These methods ignore the visual information of images, instead, they focus solely on modeling users by discovering user profiles and behavior patterns. The representations of images and users are still isolated due to semantic gap and the sparsity of user-image interactions.

Only a few recent work is concentrated on joint modeling of users and images for making recommendations~\cite{NewSIDL,SIDL,RecInfluence}. Sang \etal~\cite{RecInfluence} propose a topic sensitive model that concerns user preferences and user uploaded images to study users' influences in social networks. Liu \etal~\cite{SIDL} propose to recommend images by voting strategy according to learnt social embedded image representations. Till now, the existing methods often perform separate processing of user information and image and then simply combining them. A fully integrated solution is to be investigated.



\section{Problem Formulation of Comparative Learning}
We address the hybrid representations, \ie simultaneous representations of both users and images in a same latent space, via a deep learning approach. For this learning, how to prepare training data is not obvious. Given the fact that users rarely provide ratings on images due to the abundance of online images, we shall be able to utilize users' implicit feedback on images. Such feedback, however, is still sparse and severely unbalanced, usually negative feedback is almost none~\cite{hu2008collaborative}. A naive learning, \eg learning a distance between a user and an image, will probably fail due to the training data.

Motivated by previous efforts on deep ranking models~\cite{Sim,FineGain,Tri}, we propose a comparative learning method to tackle the imperfect training data. Several symbols are defined as follows. Let an image be $I$ and a user be $U$, we have defined functions $\pi(I)$ and $\phi(U)$ that map $I$ and $U$ to a same latent space, respectively. Another function $D$ is used to measure the distance between any two vectors in the learnt latent space. Note that instead of learning a distance between user and image, we propose to learn comparatively the relative distances between a user and two images. That is, the learning algorithm is given a sequence of triplets,
\begin{equation} \label{e1}
\{\mathcal{T}_t=(U_t, I_t^+, I_t^-), t=1,2,...,T\},
\end{equation}
where $T$ is the total amount of triplets, $U_t, I_t^+, I_t^-$ indicate the triple input elements, \ie query user $U_t$ prefers image $I_t^+$ than image $I_t^-$. Then, the learning is to find such mapping functions $\pi(\cdot)$ and $\phi(\cdot)$ and such a distance function $D(\cdot,\cdot)$, to satisfy
\begin{equation} \label{e2}
D(\pi(U_t),\phi(I_t^+)) < D(\pi(U_t),\phi(I_t^-)), \forall t.
\end{equation}



To fulfill this learning, we may perceive Eq. (\ref{e2}) as a binary classification problem (the former distance is \emph{less} or \emph{more} than the latter), and thus can reuse the 0-1 loss function, or its better alternatives such as hinge loss function. However, in order to make the distance measure more discriminative (in Eq. (\ref{e2}) the difference between the two distances should be as large as possible), we may also adopt cross entropy as loss function. Specifically, let $o_{ij}^{U_t}=D(\pi(U_t),\phi(i)) - D(\pi(U_t),\phi(j))$, and
\begin{equation} \label{e3}
\begin{split}
{P}^{U_t}_{ij} = \frac{e^{o_{ij}^{U_t}}}{1+e^{o_{ij}^{U_t}}},
\end{split}
\end{equation}
we further define
\begin{equation} \label{e4}
\begin{split}
\bar{P}^{U_t}_{ij}=\left\{
\begin{aligned}
0 & , & (i=I_t^+,j=I_t^-) \\
1 & , & (i=I_t^-,j=I_t^+)
\end{aligned}
\right.
\end{split}
,
\end{equation}
then our learning objective is defined by cross entropy as,
\begin{equation} \label{e5}
\begin{split}
\min_{\pi,\phi,D}&\mathcal{L}(\{\mathcal{T}_t\}) = \\
&\sum_t {-\bar{P}^{U_t}_{ij}\log({P}^{U_t}_{ij}) - (1-\bar{P}^{U_t}_{ij})\log(1-{P}^{U_t}_{ij})}.
\end{split}
\end{equation}

\begin{figure*}
\centerline{\includegraphics[width=\textwidth]{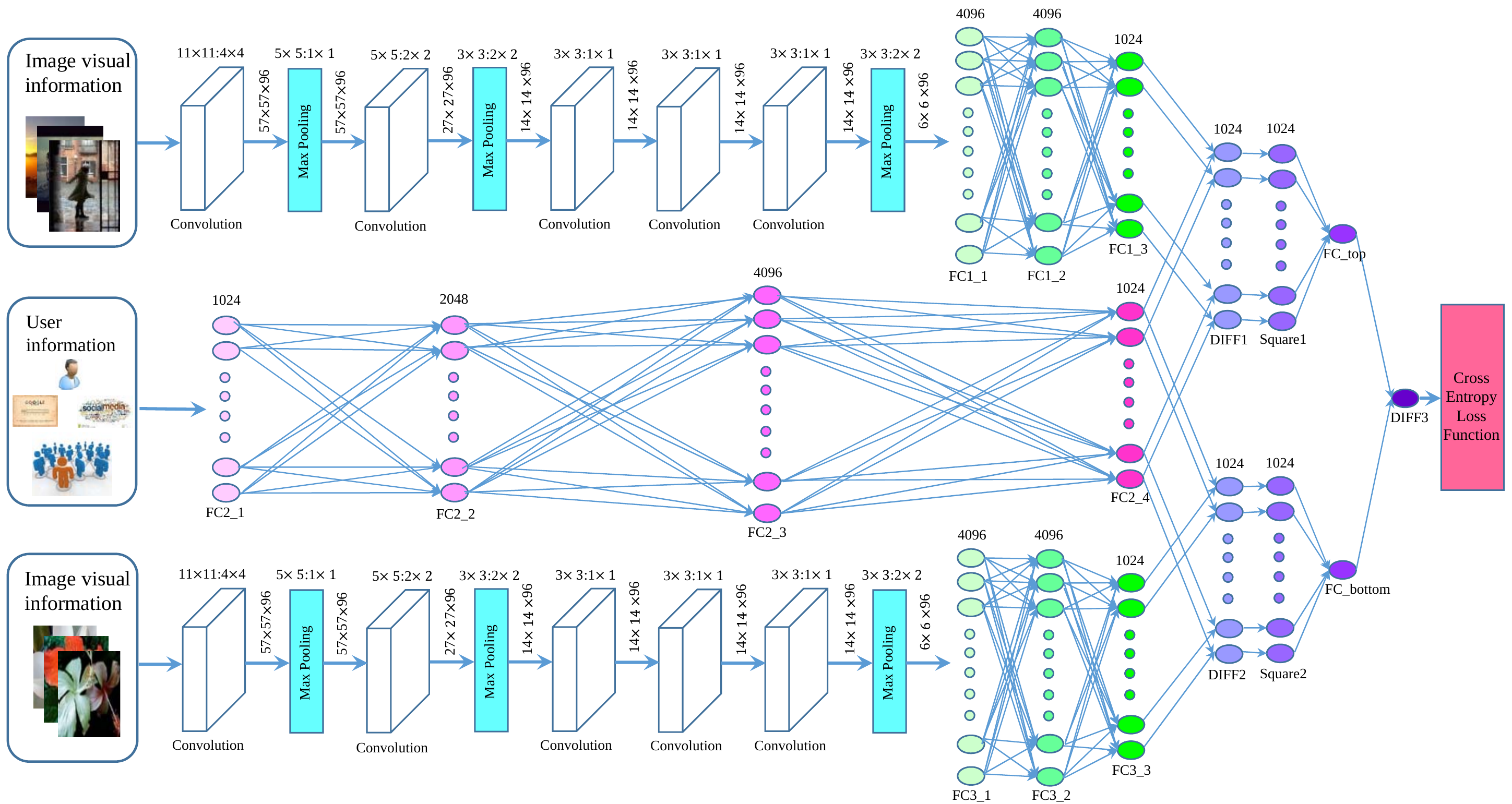}}
\caption{This figure depicts the deep network used for comparative deep learning (CDL). There are three sub-networks that all output $1024$-dim vectors as representations of images and users, respectively. The top and bottom sub-networks processing images are identical. The middle sub-network is processing users. Following these sub-networks are two distance calculating nets. The difference between distances is fed into the final cross-entropy loss function for comparison with label. The numbers shown above each arrow give the size of the corresponding output. The numbers shown above each box indicate the size of kernel and size of stride for the corresponding layer.
}\label{f2}
\end{figure*}

In this paper, we are interested in learning representations of users and images and thus we may assume the distance function $D$ to be quite simple, for example the Euclidean. Then, the comparative learning leads to solutions to the mapping functions $\pi(\cdot)$ and $\phi(\cdot)$ that generate representations seamlessly. Traditionally, such learning problems were solved by hand-crafted shallow models, but our case raises more difficulties, since it is required to learn two mapping functions at the same time and the two functions are dealing with quite different modalities but shall embed into a same space. We turn to deep learning to solve this problem.

\section{Comparative Deep Learning (CDL)}

As illustrated in Fig.~\ref{f2}, we design a deep network to perform the proposed comparative deep learning (CDL). This network architecture takes triplets as inputs, \ie $(U_t, I_t^+, I_t^-)$ with a query user $U_t$ having relatively shorter distance from image $I^+$ than from image $I^-$. There are three sub-networks in the CDL architecture. The top and bottom sub-networks are two convolutional neural networks (CNNs) with identical configuration and shared parameters, they are designed to capture image visual information. The middle sub-network is a full-connection neural network that is designed for user's information.

The two kinds of sub-networks in our architecture correspond to mapping functions for image $I:\pi(I) \in \mathcal{R}^d$ and for user $U:\phi(U) \in \mathcal{R}^d$, respectively, where $\mathcal{R}^d$ is the target latent space. The outputs of these sub-networks are indeed hybrid representations of images and users (FC1\_3, FC2\_4 and FC3\_3 in Fig. \ref{f2}). To guarantee that the learnt functions $\pi(\cdot)$ and $\phi(\cdot)$ can embed multimodal information into the same latent space, we add two distance calculating nets that outputs two distances (FC\_top and FC\_bottom in Fig. \ref{f2}), and the difference between distances, \ie $o_{ij}^{U_t}$ in Eq. (\ref{e3}), is fed into the final cross entropy loss function to be verified by the label. In the rest of this section, we will describe each part of the architecture in more details.

In the top/bottom sub-network, there are 5 convolutional layers, 3 max-pooling layers and 3 full-connection layers. These configurations including the sizes of convolution kernels in the convolution layers and the numbers of neurons in the full-connection layers are remarked in Fig.~\ref{f2}. The architecture and settings of this sub-network are inspired by AlexNet~\cite{AlexNet}, which achieves great success in modeling image visual information. Input to this sub-network are the pixel data of RGB channels of an image, and output of this sub-network is a $1024$-dim vector (FC1\_3 and FC3\_3).

The middle sub-network is designed for capturing user's information. Users' preferences/intents can be described in various forms and different kinds of data. However, normally neural networks accept only numerical vector inputs. We adopt a traditional full-connection network to map an input user vector to the representation, and leave the process of converting practical data into user vectors to be defined in Section 5. This sub-network also outputs $1024$-dim vectors (FC2\_4) to be comparable with the image representations.


Afterwards, the deep network performs distance calculation. As the focus of this paper is on effective hybrid representations, we assume the distance function shall be quite simple, yet we still design a sub-network for calculating distance. It is completed by first calculating the element-wise difference vector (DIFF1 and DIFF2 shown in Fig.~\ref{f2}), then calculating the element-wise square (Square1 and Square2 in Fig.~\ref{f2}), and finally using a full-connection layer to derive the distance. A special note is that we adopt the idea of dropout (at rate 0.5) to bring in some randomization factors to select partial dimensions of the learnt representations. The full-connection layer acts as weighting factors on the different dimensions of squared difference vector, and thus the distance calculating sub-network is equivalent to weighted $l_2$-norm distance function. Many complicated distance calculating networks can be adopted herein, but we leave them for future exploration.

\section{CDL for Image Recommendations}
Since our proposed CDL learns hybrid representations, it is well suitable for user-centric image processing tasks. In this paper, we take personalized image recommendation task as an example to discuss on the utility of CDL. We will restrict our discussions to recommending new images to a user based on her browsing history and will not dive into details of practice. There are two key issues to be solved before applying the CDL. First, how to preprocess user data to generate user vectors as inputs to the deep network. Second, how to prepare triplets as training data.



\begin{figure}[t]
\centerline{\includegraphics[width=\columnwidth]{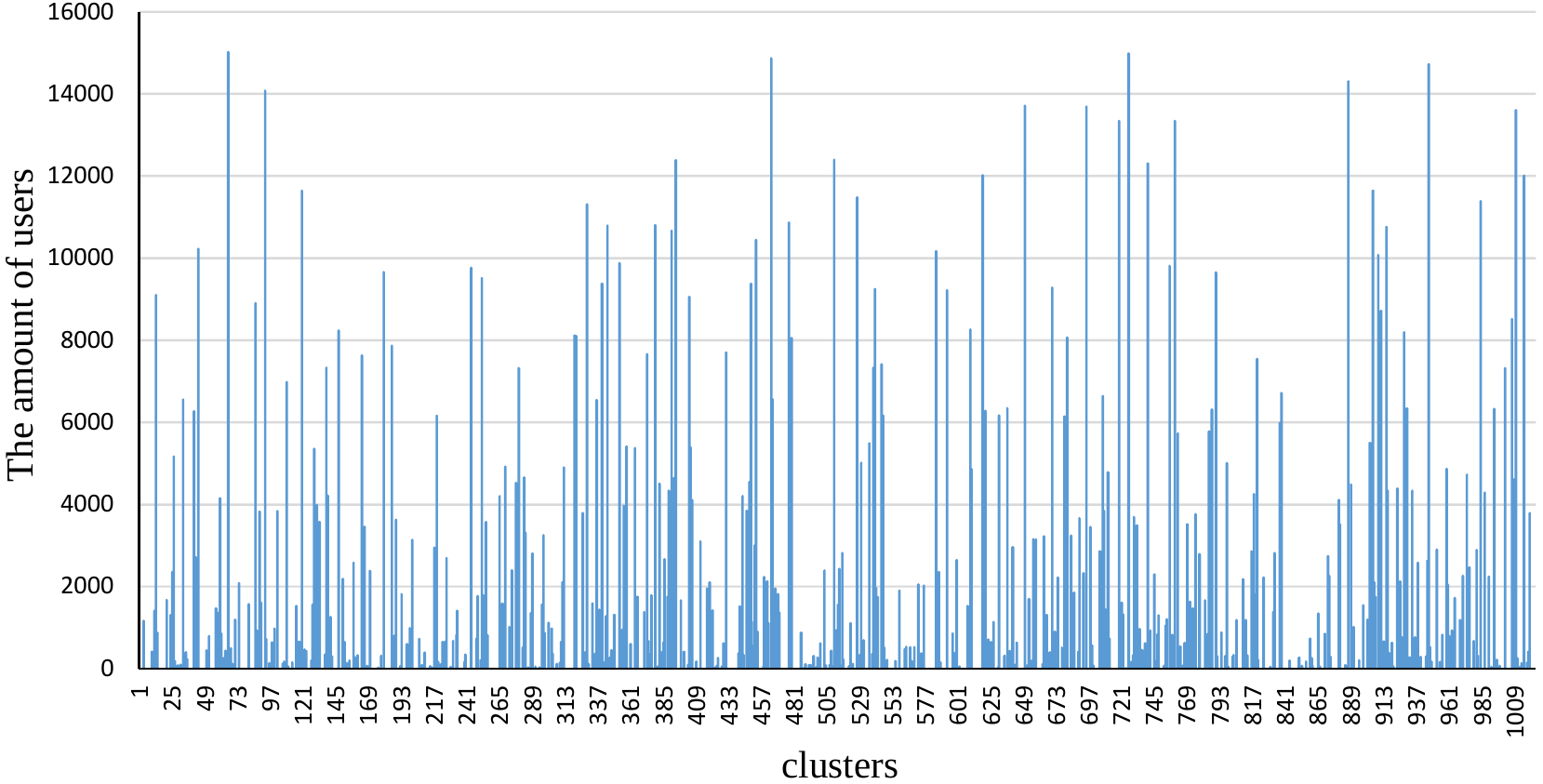}}
\caption{This figure shows the distribution of clusters. The $x$-axis displays $1024$ clusters and the $y$-axis is the number of users having interests in this cluster. A user can be described by bag of words where words are indeed clusters.}\label{cluster}
\end{figure}

There are several intuitive methods to generate user vectors. A straightforward method is bag-of-words, for example, using a vector whose dimension is equal to the amount of possible tags, and entries of this vector correspond to the interest levels of this user in these tags. Such interest levels can be estimated from the user's browsing history and tagging history, and so on. This method faces two challenges. First, tags may be too many and accordingly the vector may be too sparse. Second, the method cannot deal with synonyms of tags. In this paper, we use the well-known word2vector~\cite{w2v} as a remedy for these problems. Tags are converted to vectors\footnote{Actually we use Google trained vectors downloaded from https://code.google.com/p/word2vec/.} and then vectors are clustered by $k$-means into $1024$ semantic clusters. Then, tags are replaced by clusters and the bag-of-words method works on these clusters. Fig.~\ref{cluster} shows the distribution of clusters, where we observe the clusters have variant frequencies and bear topical polymerism to some degree.

Since the input is a set of triplets in our proposed CDL, it is desirable to generate a set of pairwise images (a positive image and a negative image) for each user. Positive images for users are often handy since users' behavior data such as ``add to favorites'' and ``like'' give such information explicitly. However, negative images are not obvious~\cite{hu2008collaborative}. An image is not ``liked'' by a user dose not necessarily indicate the user is not interested in the image, but rather the user never saw it. We utilize social data to help solve this problem. In general, a user has friendship with another usually indicates that both users have similar interests, and users of the same social group have similar interests also. For a specific user, we define the set of potentially liked images as her friends' favorite images and the images ``liked'' by users in her joined groups. We then assign the images to be negative, which have no tag of the user's interests and are not belonging to the set of potentially liked images. Due to abundance of negative images assigned in this manner, random sampling can be performed to generate a subset of triplets for training purpose.

Last but not the least question is how to make recommendations for users. This is performed in the following steps. First, a set of candidate images are selected, where each candidate shall have at least one tag of the user's interests. Second, the representations of these candidate images as well as of the user are calculated; these representations can be calculated and stored in advance, or can be calculated in parallel to accelerate. Third, distances are calculated among the images and the user. Finally, K nearest neighboring images having minimum distances are chosen as recommendations.


\begin{figure}[t]
\centerline{\includegraphics[width=\columnwidth]{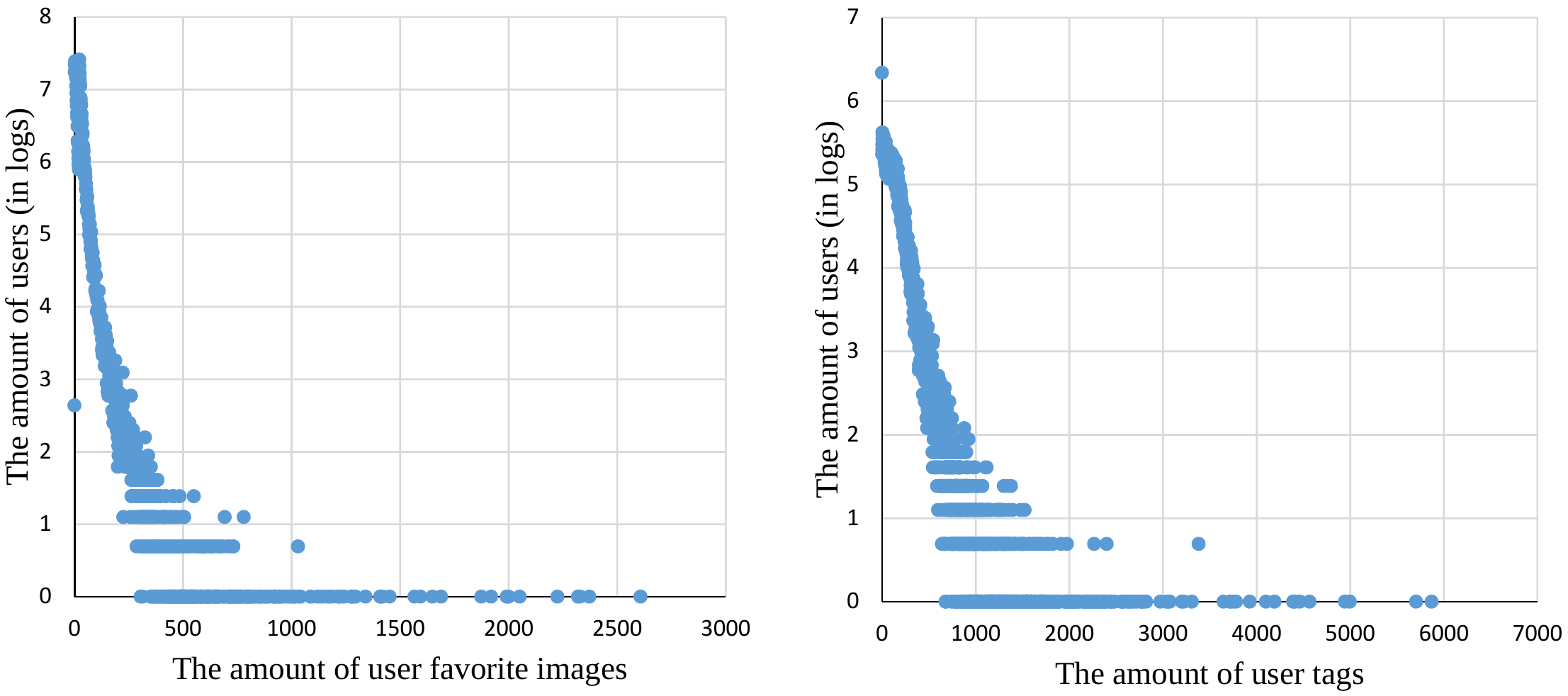}}
\caption{Distributions of the amounts of user favorite images and user tags. Note the logarithm scale of $y$-axis.}\label{distribution1}
\vspace{-12pt}
\end{figure}

\section{Experiments}

In this section, we report the conducted experiments to evaluate the efficacy of the proposed CDL of hybrid representations. We first introduce our experimental settings. Then, we evaluate the performance of our learnt hybrid representations in personalized image recommendation task. Finally, we give some insights of our proposed approach.

\subsection{Experimental Settings}
\subsubsection{Datasets}

In this paper, we use the same dataset as reported in~\cite{SIDL}. The images and users' information in this dataset are crawled from Flickr through its API. There are $101,496$ images, $54,173$ users, $6,439$ groups and $35,844$ tags in this dataset. The details of crawling can be found in~\cite{SIDL}. On average, there are $23.5$ tags and $5.8$ favorite images for each user. Due to the sparsity of user-image interactions, this dataset is not quite suitable for traditional recommendation algorithms, especially collaborative filtering. Therefore, we do not compare our method with them.

The distributions of the amounts of user favorite images and user tags are shown in Fig.~\ref{distribution1}. Both of them are typical long-tail distributions. Users having modest favorite images and tags usually have most valuable and robust information~\cite{long}. Too few favorites indicate inactivity of user, and too many favorites indicate quite diverse interests of user. Thus, we filter out users that have less than $40$ or more than $200$ favorite images from test~\cite{SIDL}. Furthermore, according to statistics shown in Fig.~\ref{cluster}, we further filter out users that have interests in less than $80$ or more than $280$ clusters from training data, so as to improve the accuracy of training, but keep them for test. Finally, we have $8,616$ users for training and $15,023$ users for test.

For each user, $20$ images are randomly selected from her favorite images and ``concealed'' for test. Training data are then generated by randomly sampling the rest favorite images as well as assigned negative images (c.f. Section 5). $20$ triplets are sampled for each user for training. Finally, there are $72,161$ distinct images in training data. After training, the concealed favorite images are retrieved and mixed with other $80$ images (for each user) for test.

\subsubsection{Compared Approaches}
As discussed in Section 3, our proposed CDL allows the choice of different loss functions and we have used cross entropy for better performance. We also tested the use of hinge loss in replacement of cross entropy. Moreover, we compare our method with several state-of-the-art approaches.

\textbf{Borda Count with SIDL~\cite{Borda,SIDL}.} Social embedding Image Distance Learning (SIDL) is a novel image distance learning method that embeds the similarity of collective social and behavioral information into visual space. After learning the social embedding image distance metric, it can be adopted together with Borda Count method~\cite{Borda} to perform personalized image recommendations, as detailed in~\cite{SIDL}.

\textbf{Borda Count with BoW, ImageNet~\cite{AlexNet}, LMNN~\cite{LMNN}, Social+LMNN~\cite{SIDL}.} Bag of Words (BoW) feature is a traditional hand-crafted visual representation, and ImageNet feature stands for deep learning based representation, both can be used to measure image similarity with \eg Euclidean distance. Large Margin Nearest Neighbor (LMNN) is a metric learning method to reduce the margins of the nearest neighbors. Liu \etal proposes to embed social similarity into LMNN, termed Social+LMNN. We then use BoW, ImageNet, LMNN, and Social+LMNN with Borda Count method to evaluate the performance of personalized image recommendations.

\textbf{TwoNets.} In this paper, we propose the CDL instead of naive deep learning to learn hybrid representations. To demonstrate the effectiveness of CDL, we also perform the naive learning experiment called TwoNets. Specifically, TwoNets is similar to the CDL but it has only two sub-networks, which process user and image, respectively; the output representations of the two sub-networks are directly compared to calculate a distance, and the distance is re-scaled by a logistic sigmoid function and then compared with the ground-truth by cross entropy loss function; note that in TwoNets, training data consist of doublets of $(user, image)$ and the ground-truth is 0 or 1 indicating negative or positive.

\subsubsection{Implementation}
We implement the CDL and TwoNets methods based on the open source deep learning software Caffe~\cite{Caffe}. In our experiments, all images are resized to $256\times256$. The structure and parameters of sub-networks are illustrated in Fig.~\ref{f2}, and all probabilities of dropout are set to 0.5~\cite{AlexNet}. The learning rate starts from 0.001 for all layers and the momentum is 0.9. The mini-batch size of images is 128. The weight decay parameter is 0.0005. Training was done on a single GeForce Tesla K20c GPU with 5GB graphical memory, and it took about 4 days to finish training.

\subsection{Overall Performance}
\begin{figure}[t]
\centerline{\includegraphics[width=0.4\textwidth]{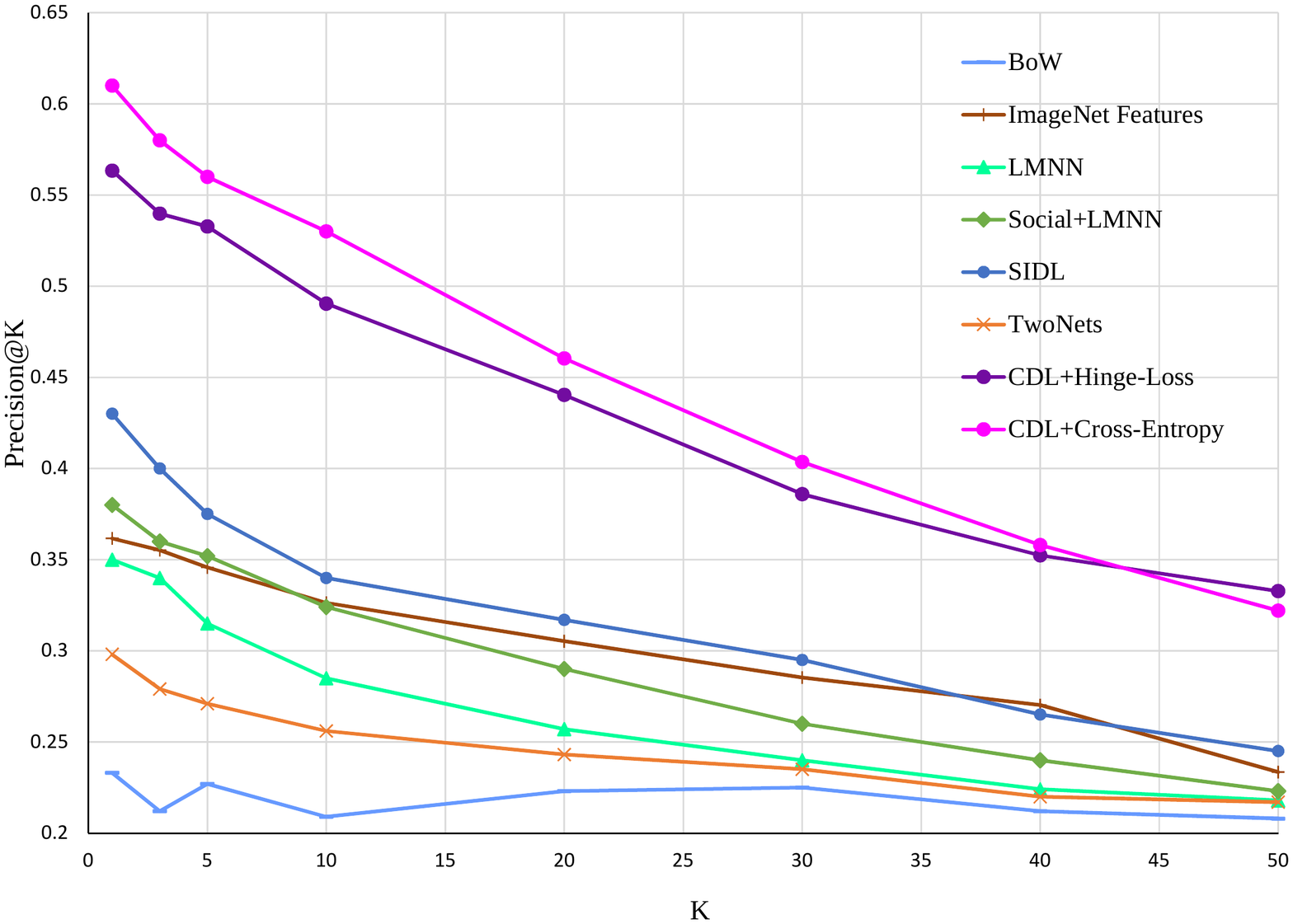}}
\caption{Precision@K for different K values of compared image recommendation methods.}\label{precision}
\end{figure}

\begin{figure}[t]
\centerline{\includegraphics[width=0.4\textwidth]{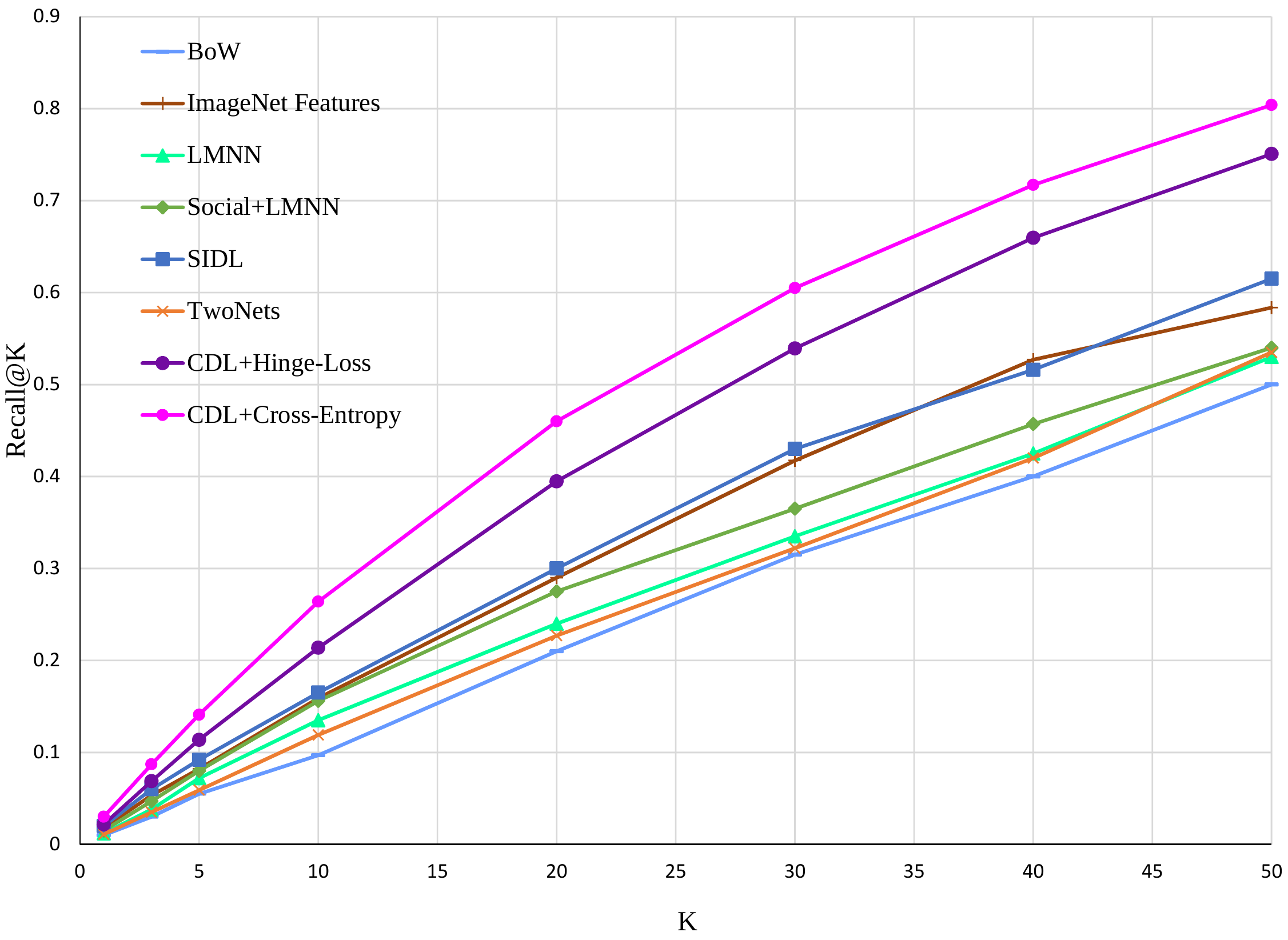}}
\caption{Recall@K for different K values of compared image recommendation methods.}\label{Recall}
\end{figure}

In our personalized image recommendation task, the target is to recommend 20 images out of 100 candidates for each user. To make a fair comparison, we implement every comparative method to return top K recommended images where K is adjustable. Precision@K and Recall@K are used to evaluate the performance of each method, which are shown in Figs.~\ref{precision} and~\ref{Recall}, respectively. It can be seen that our approach performs the best in both precision and recall for all K values, which demonstrates the effectiveness of our proposed CDL of hybrid representations. Note that using cross entropy as loss function has obvious advantage compared to using hinge loss function in our image recommendation task (c.f. Section 3).

The approaches based solely on hand-crafted visual representations, \ie BoW and LMNN, perform poorly in making recommendations. The Precision result of BoW is near to random guess (random guess for recommending 20 out of 100 achieves precision 0.2). ImageNet Features lead to much better results, almost the third best after our CDL methods and SIDL, which shows the advantage of deep learning based representations.

If we add social factors to constrain LMNN (\ie Social+LMNN), the performance will be improved a lot, due to the utilization of extra information besides visual features. The SIDL performs better than Social+LMNN, indicating the importance of carefully designed features to capture visual information and embedding functions to integrate multimodal information. Compared to SIDL, our approach leads significant gains of average $42.58\%$ and $46.50\%$ for precision and recall, respectively. It owes to the superiority of deep network models over traditional hand-crafted models especially in capturing visual information.


It should be noted that TwoNets, also adopting deep network model, has very poor performance. It is slightly better than BoW but the latter is near to random guess. Thus, deep network models do not guarantee great success especially when the task is complicated (learning hybrid representations) and the training data are imperfect (unreliable negative samples). The proposed CDL outperforms TwoNets significantly and consistently, which further demonstrates the effectiveness of the proposed comparative learning method.

\begin{figure*}
\centerline{\includegraphics[width=\textwidth]{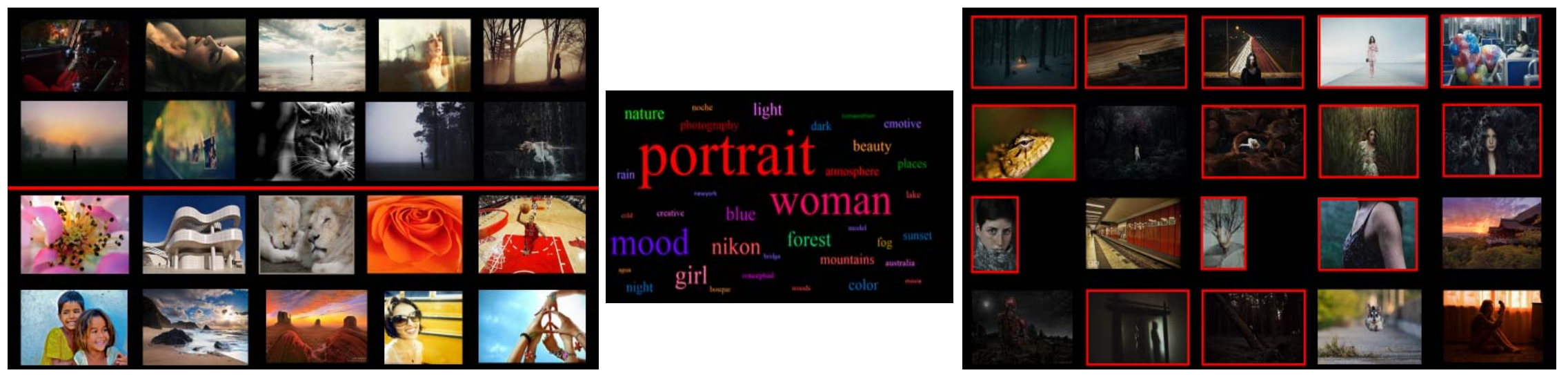}}
\caption{(Best view in color.) Case study of making recommendations to a selected user. Left: some samples of training images for this user, 10 positive and 10 negative, separated by the red line; unlike positive images that are indeed favorite images of this user, negative images are ``assigned'' by the process discussed in Section 5. Middle: the word cloud of this user's frequent tags retrieved from her tagging history and browsing history. Right: recommendation results sorted in relevance (ascending order of distance calculated by hybrid representations), where correct results are highlighted by red borders.}\label{casestudy}
\end{figure*}

%

\begin{figure}[t]
\centerline{\includegraphics[width=\columnwidth]{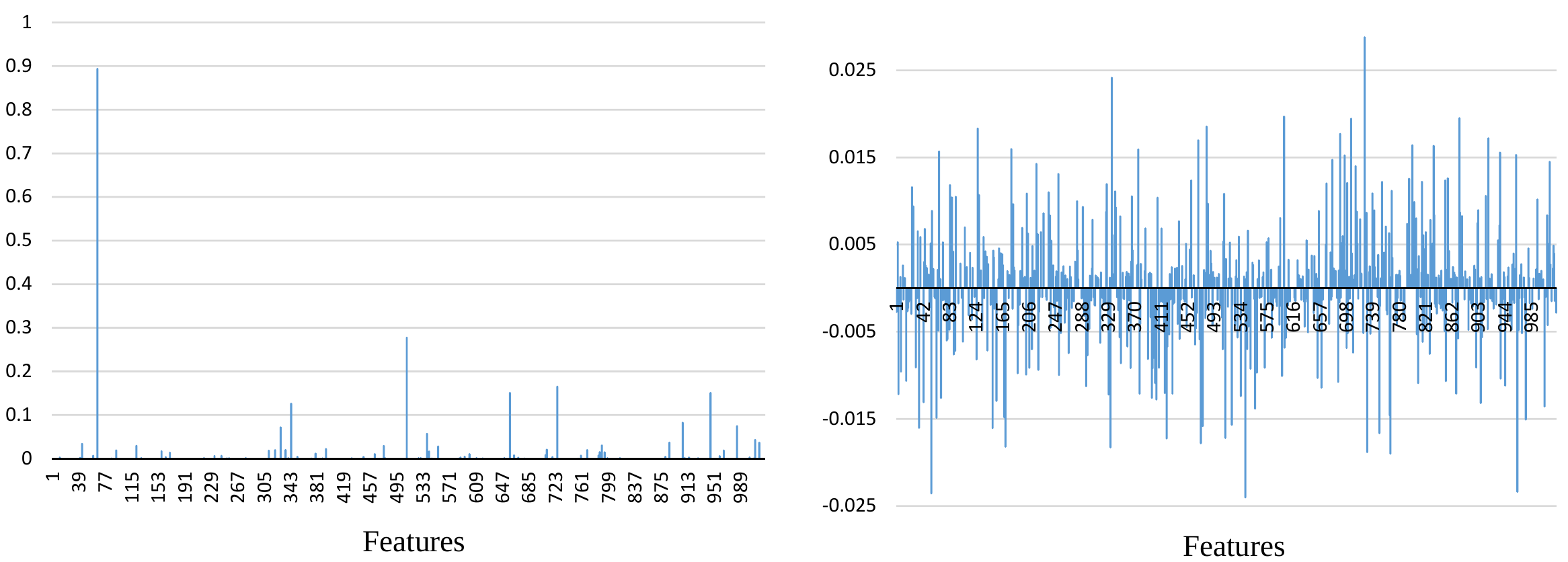}}
\caption{Exemplar input and output of the user sub-network in our designed dual-net deep network. Left: pre-processed user vector (input). Right: learnt user representation (output).}\label{features}
\end{figure}

\subsection{Case Study and Insights}

In this section, we present a case for comprehensive study to give some insights of our proposed approach. For the selected user whose word cloud of frequent tags can be found in Fig.~\ref{casestudy} (Middle), we prepare a set of images for training, illustrated in Fig.~\ref{casestudy} (Left). It can be observed that positive images match the user's preferences as described by the word cloud, \eg portrait, woman and mood. Obviously, there are large differences between positive images and negative images, which verifies the effectiveness of our designed process for assigning negative images for training (c.f. Section 5).

Our approach's recommendation results for this user are shown in Fig.~\ref{casestudy} (Right). Precision@20 is as high as $70\%$ in this case. Given a closeup view, most of correct recommendations made by our approach are portraits with darker tone and gloomy atmosphere, in the similar topics and styles of the user's word cloud and the positive images in training data. Interestingly, the 1st image in the 2nd row in Fig.~\ref{casestudy} (Right) is not belonging to the styles mentioned above. Such images are not easy to be recommended if using purely tags. But we may compare this image with the 2nd image in the 2nd row in Fig.~\ref{casestudy} (Left), and observe their similarity in the sense of color, bokeh, and theme. This is where image representations help.

\begin{figure}[t]
\centerline{\includegraphics[width=\columnwidth]{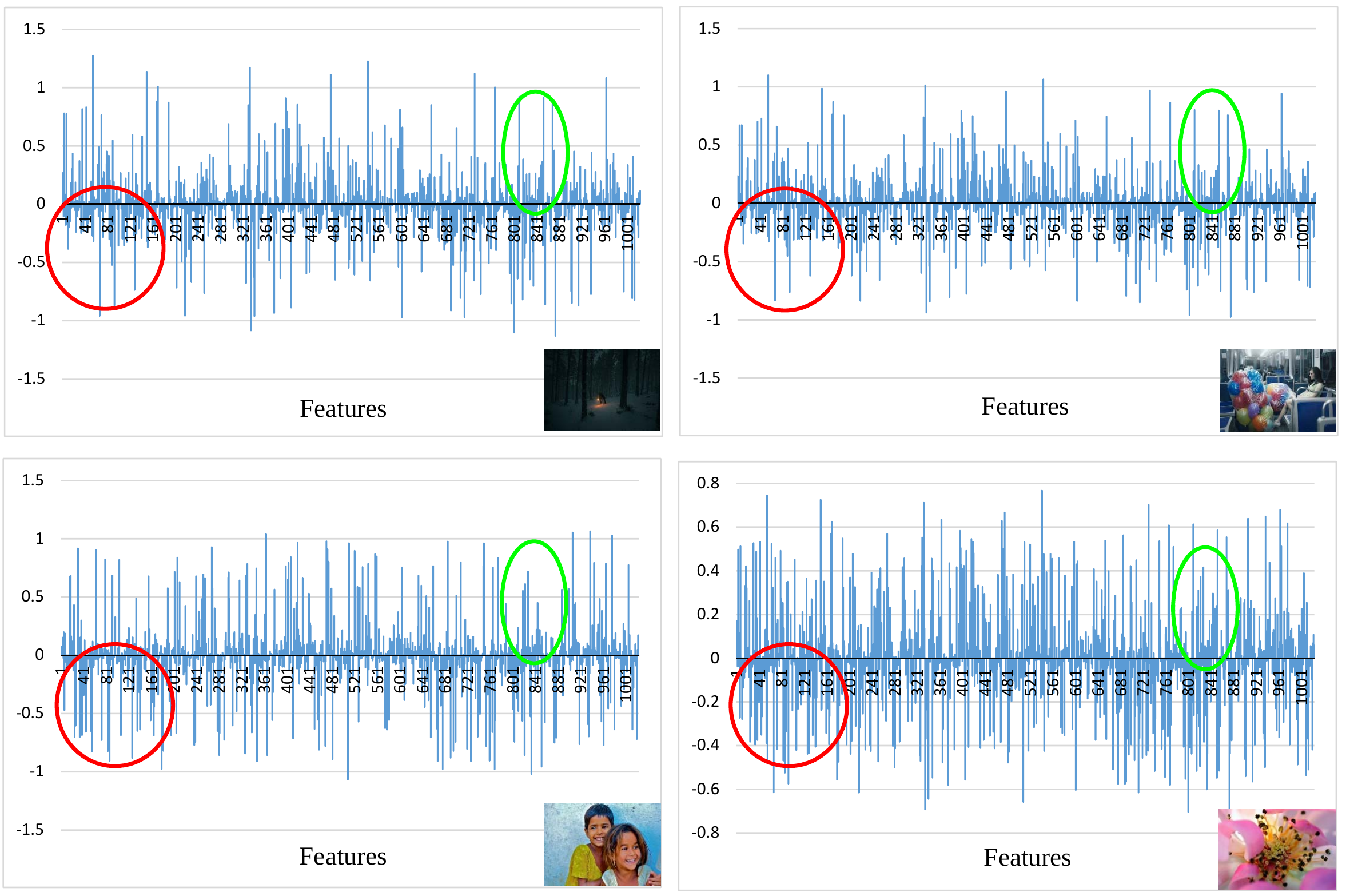}}
\caption{Exemplar learnt representations of positive images (top row) and negative images (bottom row). Note the similarity between positive images and dissimilarity between positive and negative images, especially in the circled areas.}\label{imgfea}
\end{figure}

Several images in the recommendation results are not ``correct'' according to ground-truth, but we cannot say firmly that the user dislikes these images since we do not know whether the user has ever seen them. Especially, the 2nd image in the 2nd row and the last image in Fig.~\ref{casestudy} (Right) are probably what user may like. Both images match the word cloud and the user's positive images in the training data. However, the 1st image in the 4th row is probably a mistake of recommendation. This photo has darker background and a human-like object (which is actually a skeleton), which interprets its being selected, but per view of the user's positive images, skeleton may not be his/her favorite. In such cases, making finer discrimination of similar objects may help improve the recommendation accuracy.

Fig.~\ref{features} illustrates the input and output of the user sub-network learnt by CDL. Fig.~\ref{features} (Left) is the input vector, indeed a bag-of-words vector spanned over clusters of word2vector results, such vector is very sparse, dominated by several interests. Fig.~\ref{features} (Right) is the learnt user representation, not sparse any more. It shall be noted that such dense vectors are due to the following distance calculation (weighted $l_2$-norm distance).

Fig.~\ref{imgfea} illustrates several examples of learnt image representations, where the top row shows positive images and the bottom row shows negative images. It is interesting to find that the representations of two positive images are quite similar, while they are very different from the representations of two negative images. Some obvious similarity and dissimilarity are highlighted by circles in the figure. However, such information is not easily perceived from the images themselves. Therefore, simultaneous learning of hybrid representations is indeed quite different from only learning the representations of images.


\section{Conclusions}

In this paper, we explore learning of hybrid representations to capture both visual information and intents or preferences of users over images, and utilizing such representations for user-centric tasks such as personalized image recommendations. A dual-net deep network model is proposed to learn representations in a latent semantic space. We also propose a comparative deep learning method to train the designed deep network, in which triplets of users and positive/negative images are taken as inputs and the relative distances are the objective of learning. The empirical evaluations on personalized image recommendation task show that our proposed approach achieves much better performance than naive deep learning as well as several state-of-the-art image recommendation solutions. The proposed comparative deep learning can be applied to many other user-centric applications, such as image search and image editing. We will further explore along these directions.

{\small
\bibliographystyle{ieee}
\bibliography{egbib}

\begin{thebibliography}{10}\itemsep=-1pt

\bibitem{adomavicius2005toward}
G.~Adomavicius and A.~Tuzhilin.
\newblock Toward the next generation of recommender systems: A survey of the
  state-of-the-art and possible extensions.
\newblock {\em TKDE}, 17(6):734--749, 2005.

\bibitem{Borda}
J.~A. Aslam and M.~Montague.
\newblock Models for metasearch.
\newblock In {\em SIGIR}, pages 276--284, 2001.

\bibitem{TraRec}
P.~Cui, Z.~Wang, and Z.~Su.
\newblock What videos are similar with you?: Learning a common attributed
  representation for video recommendation.
\newblock In {\em ACM Multimedia}, pages 597--606, 2014.

\bibitem{HOG}
N.~Dalal and B.~Triggs.
\newblock Histograms of oriented gradients for human detection.
\newblock In {\em CVPR}, pages 886--893, 2005.

\bibitem{TraRec3}
J.~Fan, D.~A. Keim, Y.~Gao, and H.~Luo.
\newblock Justclick: Personalized image recommendation via exploratory search
  from large-scale flickr images.
\newblock {\em TCSVT}, 19(2):1051--8215, 2009.

\bibitem{hu2008collaborative}
Y.~Hu, Y.~Koren, and C.~Volinsky.
\newblock Collaborative filtering for implicit feedback datasets.
\newblock In {\em ICDM}, pages 263--272, 2008.

\bibitem{Hua}
X.-S. Hua, L.~Yang, J.~Wang, J.~Wang, M.~Ye, K.~Wang, Y.~Rui, and J.~Li.
\newblock Clickage: towards bridging semantic and intent gaps via mining click
  logs of search engines.
\newblock In {\em ACM Multimedia}, pages 243--252, 2013.

\bibitem{SocialRec2}
M.~Jamali and M.~Ester.
\newblock Trustwalker: a random walk model for combining trust-based and
  item-based recommendation.
\newblock In {\em KDD}, pages 397--406, 2009.

\bibitem{Caffe}
Y.~Jia, E.~Shelhamer, J.~Donahue, S.~Karayev, J.~Long, R.~Girshick,
  S.~Guadarrama, and T.~Darrell.
\newblock Caffe: Convolutional architecture for fast feature embedding.
\newblock arXiv:1408.5093, 2014.

\bibitem{SocialRec1}
M.~Jiang, P.~Cui, F.~Wang, W.~Zhu, and S.~Yang.
\newblock Scalable recommendation with social contextual information.
\newblock {\em TKDE}, 26(11):2789--2802, 2014.

\bibitem{TraRec2}
Y.~Jing, X.~Zhang, L.~Wu, and J.~Wang.
\newblock Recommendation on flickr by combining community user ratings and item
  importance.
\newblock In {\em ICME}, pages 1--6, 2014.

\bibitem{MF}
Y.~Koren, R.~Bell, and C.~Volinsky.
\newblock Matrix factorization techniques for recommender systems.
\newblock {\em Computer}, 48:30--37, 2009.

\bibitem{AlexNet}
A.~Krizhevsky, I.~Sutskever, and G.Hinton.
\newblock Imagenet classification with deep convolutional neural networks.
\newblock In {\em NIPS}, pages 1106--1114, 2012.

\bibitem{Sim}
H.~Lai, Y.~Pan, Y.~Liu, and S.~Yan.
\newblock Simultaneous feature learning and hash coding with deep neural
  networks.
\newblock In {\em CVPR}, pages 3270--3278, 2015.

\bibitem{repreExam1}
K.~Lenc and A.~Vedaldi.
\newblock Understanding image representations by measuring their equivariance
  and equivalence.
\newblock In {\em CVPR}, pages 991--999, 2015.

\bibitem{RecIcme}
Y.~Li, J.~Luo, and T.~Mei.
\newblock Personalized image recommendation for web search engine users.
\newblock In {\em ICME}, pages 1--6, 2014.

\bibitem{repreExam5}
T.~Lin, Y.~Cui, S.~Belongie, and J.~Hays.
\newblock Learning deep representations for ground-to-aerial geolocalization.
\newblock In {\em CVPR}, pages 5007--5015, 2015.

\bibitem{NewSIDL}
S.~Liu, P.~Cui, W.~Zhu, and S.~Yang.
\newblock Learning socially embeded visual representation from scratch.
\newblock In {\em ACM Multimedia}, pages 109--118, 2015.

\bibitem{SIDL}
S.~Liu, P.~Cui, W.~Zhu, S.~Yang, and Q.~Tian.
\newblock Social embedding image distance learning.
\newblock In {\em ACM Multimedia}, pages 617--626, 2014.

\bibitem{SIFT}
D.~G. Lowe.
\newblock Object recognition from local scale-invariant features.
\newblock In {\em ICCV}, pages 1150--1157, 1999.

\bibitem{repreExam6}
A.~Mahendran and A.~Vedaldi.
\newblock Understanding deep image representations by inverting them.
\newblock In {\em CVPR}, pages 5188--5196, 2015.

\bibitem{w2v}
T.~Mikolov, K.~Chen, G.~Corrado, and J.~Dean.
\newblock Efficient estimation of word representations in vector space.
\newblock In {\em ICLR Workshop}, 2013.

\bibitem{Pan}
Y.~Pan, T.~Yao, T.~Mei, H.~Li, C.~W. Ngo, and Y.~Rui.
\newblock Click-through-based cross-view learning for image search.
\newblock In {\em SIGIR}, pages 717--726, 2014.

\bibitem{ContentFilter}
M.~J. Pazzani.
\newblock A framework for collaborative, content-based and demographic
  filtering.
\newblock {\em Artificial Intelligence Review}, 13:393--408, 1999.

\bibitem{RecInfluence}
J.~Sang and C.~Xu.
\newblock Right buddy makes the difference: An early exploration of social
  relation analysis in multimedia applications.
\newblock In {\em ACM Multimedia}, pages 19--28, 2013.

\bibitem{long}
B.~Sigurbj{\"o}rnsson and R.~Van~Zwol.
\newblock Flickr tag recommendation based on collective knowledge.
\newblock In {\em WWW}, pages 327--336, 2008.

\bibitem{FineGain}
J.~Wang, Y.~Song, T.~Leung, C.~Rosenberg, and J.~Wang.
\newblock Learning fine-grained image similarity with deep ranking.
\newblock In {\em CVPR}, pages 1386--1393, 2014.

\bibitem{LLC}
J.~Wang, J.~Yang, K.~Yu, F.~Lv, T.~Huang, and Y.~Gong.
\newblock Locality-constrained linear coding for image classification.
\newblock In {\em CVPR}, pages 3360--3367, 2010.

\bibitem{LMNN}
K.~Q. Weinberger and L.~K. Saul.
\newblock Distance metric learning for large margin nearest neighbor
  classification.
\newblock {\em JMLR}, 10:207--244, 2009.

\bibitem{Tri}
P.~Wu, S.~C. Hoi, H.~Xia, P.~Zhao, D.~Wang, and C.~Miao.
\newblock Online multimodal deep similarity learning with application to image
  retrieval.
\newblock In {\em ACM Multimedia}, pages 153--162, 2013.

\bibitem{repreExam4}
R.~Xia, Y.~Pan, H.~Lai, C.~Liu, and S.~Yan.
\newblock Supervised hashing for image retrieval via image representation
  learning.
\newblock In {\em AAAI}, pages 2156--2162, 2014.

\bibitem{repreExam3}
Z.~Xu, Y.~Yang, and A.~G. Hauptmann.
\newblock A discriminative cnn video representation for event detection.
\newblock In {\em CVPR}, pages 1798--1807, 2015.

\bibitem{LatentFea}
Z.~Yuan, J.~Sang, Y.~Liu, and C.~Xu.
\newblock Latent feature learning in social media network.
\newblock In {\em ACM Multimedia}, pages 253--262, 2013.

\bibitem{repreExam2}
M.~D. Zeiler and R.~Fergus.
\newblock Visualizing and understanding convolutional networks.
\newblock In {\em ECCV}, pages 818--833, 2014.

\end{thebibliography}
}

\end{document}